\documentclass[11pt]{article}

\usepackage{graphicx}
\usepackage{amsmath,amssymb,amsfonts}
\usepackage{mathrsfs}
\usepackage{xcolor}
\usepackage{textcomp}
\usepackage{booktabs}
\usepackage{algorithm}
\usepackage{algorithmicx}
\usepackage{algpseudocode}
\usepackage{listings}
\usepackage{geometry}
\usepackage{hyperref}
\usepackage{float}
\usepackage{caption}
\usepackage{subcaption}
\usepackage{multirow}

\hypersetup{
    colorlinks=true,
    linkcolor=blue,
    citecolor=blue,
    urlcolor=blue
}

\makeatletter
\def\@makefnmark{\hbox{\@textsuperscript{\normalfont\@thefnmark}}}
\makeatother

\begin{document}

\title{Improvement of Robot's Simultaneous Localization and Mapping Using an Effective Transformation to Achieve Linear Model}

\author{
    Seyed Farzad Bahreinian$^{1}$\thanks{f.bahr@iut.ac.ir} \and
    Maziar Palhang$^{2,*}$\thanks{palhang@iut.ac.ir} \and
    Mohammad Reza Taban$^{2}$\thanks{mrtaban@iut.ac.ir} \and
    Hasan Enami Eraghi$^{2}$\thanks{hasan.enami2015@gmail.com}
}

\footnotetext[1]{Subsea Science and Technology Research Institute, Isfahan University of Technology, Isfahan, Iran}
\footnotetext[2]{Department of Electrical and Computer Engineering, Isfahan University of Technology, Isfahan, Iran}

\date{}
\maketitle

\begin{abstract}
Nowadays mobile robots have wide engineering applications. Simultaneous localization and mapping (SLAM) is an important task of these robots. The major and common algorithms used for this task are based on extended Kalman filter (EKF). One of the main problems in EKF-based SLAM is its divergence. The nonlinearity of motion and observation models and linearization error are the main reasons for the divergence. There have been some efforts to address this problem with limited success. In this paper, by applying a simple compass and using an effective transformation, we transform the non-linear state space model into a linear model. Then, by applying the original KF to this model, we reach a new method, which is called LMKF SLAM. We show that the LMKF SLAM is significantly superior to the state-of-the-art methods, especially EKF-based SLAMs, both in accuracy, convergence, and computational complexity. The proposed method is also more stable with respect to the uncertainty of sensors values and changes in system parameters. Experimental results verify these points.
\end{abstract}


\section{Introduction}\label{sec1}

Robot localization is one of the main and fundamental concerns in mobile robot navigation. When a mobile robot is applied for a mission, it must be aware of its position in the environment. The accuracy of mobile robot positioning definitely affects its performance. Certainly, without precise localization, robots may be lost and cannot be successful in their duties. In some situations, the surrounding environment of the robot is unknown too. In these cases, the robot can use simultaneous localization and mapping (SLAM) techniques. In fact, the robot gradually makes the map which is the reference for positioning of the robot \cite{Durrant2006}.

Two approaches have been proposed to implement SLAM: smoothing and filtering methods \cite{Grisetti2010, Ho2015}. The smoothing methods are frequently offline and often require a primary graph map or sub-maps to produce the optimized map \cite{Kummerle2011, Zhao2013}. The filtering methods use sensors mounted on the robot, map the environment, and localize the robot just online. Among the filtering methods, the extended Kalman filter (EKF), extended information filter (EIF), sparse EIF, Fast SLAM, and particle filter (PF) are popular \cite{Dissanayake2011, thrun2004}.

A well-known filtering method is Fast SLAM. In this algorithm, the particle filter is used to estimate the vehicle position, while the EKF is used to estimate the landmarks' positions. A development of Fast SLAM is unscented Fast SLAM (UnFS) which is based on the scaled unscented transformation (SUT) \cite{Kim2008}. The UnFS algorithm updates the mean and covariance of landmarks' state vector by the Unscented filter to avoid linearization errors and Jacobian calculations in the estimation of landmark position.

Chandra et. al proposed cubature Kalman filter (CKF) SLAM which does not require calculation of the Jacobian matrix during the prediction and update steps \cite{Chandra2011}, unlike the EKF. To overcome some drawbacks of CKF-SLAM such as low precision and poor stability, improved CKF (ICKF) SLAM has been presented [3], which uses information matrix in both prediction and update steps. The ICKF method is more accurate than the CKF-SLAM and is one of the best current filtering methods in SLAM.

During the past two decades, considerable efforts have been performed to implement SLAM and overcome its challenges such as consistency and convergence. The convergence of robot and landmarks positioning in SLAM based on KF has been proved \cite{Dissanayake2001}. However, this proof is based on a rare assumption of the linearity of robot motion and observation models. Without these assumptions, the SLAM will be inconsistent and diverges especially in large environments with a high number of landmarks. So far, few efforts have been carried out on the SLAM convergence, and among them, only a limited number of effective solutions has been proposed to improve it.

Julier et al. proved that even when the robot is stationary with no motion noise and by using bearing and range sensors, an inconsistent map will be produced by EKF-SLAM \cite{Julier2001}. This means that landmarks' positions are estimated incorrectly. This inconsistency is due to nonlinear motion and observation models. The linearization error increases the Kalman gain which then causes the estimated covariance matrix becomes less than its real value (reduction in uncertainty). Finally, this error causes incorrect state vector updating and jumps in robot and landmarks positions.

Castellanos et al. confirmed that the main reason for the inconsistency of EKF-SLAM is linearization error. They improved the consistency of EKF-SLAM by using a series of independent local maps and limiting the inconsistency level during continuous improvement of maps \cite{Castellanos2004, Castellanos2007}.

Bailey et al. showed that the EKF-SLAM inconsistency very much depends on the estimation of the robot angle \cite{Bailey2006}. If the uncertainty in this angle estimation is more than one or two degrees, inconsistency in the map and robot localization will occur. Error in the robot angle estimation leads to error accumulation, and this causes the estimated map gets away from the real map. In large environments, this problem is inevitable. The conventional solutions such as adding noise stabilizer, using iterative-EKF or Unscented KF is not able to improve the inconsistency. If the variance of the robot angle is a small value, these inconsistencies grow slowly.

Some researchers have even worked on the effects of the robot angle error estimation on upper and lower bounds of landmarks positioning error \cite{Mourikis2004, Huang2006, Huang2007}.

Huang et al. reviewed the inconsistency of the EKF-SLAM from the system observability point of view \cite{Huang2008}. Their mathematical analysis showed that the error in the linearized model of the system state has the unobservable subspace of dimension two, while the error in the nonlinear model has three unobservable degrees of freedom related to the location and the robot angle in the global reference frame. This causes the incorrect estimation of filter throughput and inappropriate reduction of covariance matrix estimations. To solve it at first, they proposed a first-estimates Jacobian EKF (FEJ-EKF) method. In this method, the first-ever available estimates for all state variables are used in EKF Jacobian \cite{Huang2009}. Then, they proposed another method called observability-constrained EKF (OC-EKF). In this approach, linearization points in EKF must be selected such that the expected errors are minimized according to observability constraints \cite{Huang2010}.

Meanwhile, some other researchers tried to investigate the effects of different parameters such as measurement noise and sampling time on EKF-SLAM consistency \cite{Hui2009, Lee2011}; nonetheless they did not propose any new approach to improve EKF-SLAM consistency or performance.

Recently, Barrau et al. have presented an improved approach in OC-EKF method based on Invariant EKF (IEKF). They have used an appropriate non-linear estimation error that causes a better consistency in the SLAM \cite{Barrau2014, Barrau2015}.

As was mentioned, a few works have been carried out to improve EKF-SLAM consistency. In all of these methods, efforts have been made to compensate for the EKF error and inconsistency caused by EKF linearization. However, in most SLAM methods, the motion and observation models of the robot are non-linear. Thus they suffer from divergence, although the divergence may occur gradually \cite{Bailey2006}.

The contribution of this paper is introducing a simple transformation to convert the non-linear state space model to a linear model. This technique only needs a simple and ordinary electronic compass. The proposed model increases the consistency and accuracy of SLAM. The convergence of this method is guaranteed as well. In comparison with the standard EKF, UnFS, and ICKF SLAM methods, the proposed LMKF SLAM has much better performance and is more stable with respect to the uncertainty of system sensors and the changes in system parameters. These indicate the significant effects of the proposed models.

The rest of the paper is organized as follows. In section 2, after introducing the primary state-space model of a common SLAM, the new/linear motion and observation models are obtained. The convergence of the new approach is also discussed. In section 3, the performance of the proposed method is evaluated in an artificial environment, and then on the Victoria Park dataset. Section 4 is dedicated to the conclusion.

\textit{Notations}: We denote matrices, vectors, and scalars by bold uppercase, bold lowercase, and italic letters, respectively. Vectors are by default in column orientation. Also, $\mathbf{I}_i$ and $\mathbf{0}_{i \times j}$ are $i\times i$ identity matrix and $i\times j$ zero matrix, respectively. The superscripts "${}^T$" and "${}^{-1}$" denote the transpose and inverse of a matrix, respectively. $|{\mathcal{I}}|$ is the cardinality of set ${\mathcal{I}}$ and $\text{diag}\{.\}$ and $\text{blkdiag}\{.\}$ denote diagonal and block diagonal matrices respectively. The symbol $\otimes$ is the Kronecker product and $\delta_j$ denotes the Kronecker delta function.

\section{Linear Model KF (LMKF) SLAM}\label{sec2}

\begin{figure}[h]
    \centering
    \includegraphics[width=6.4cm, height=5cm]{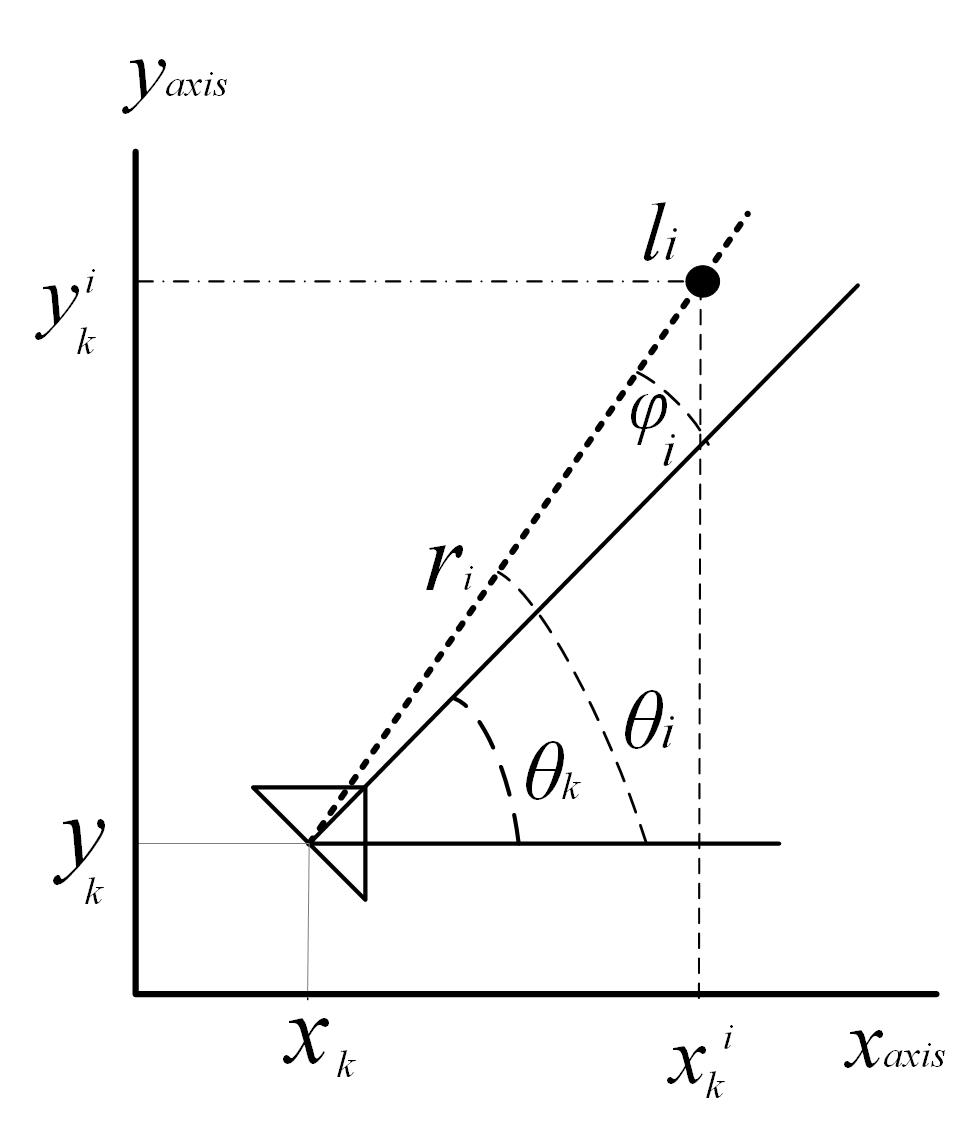}
    \caption{Robot observes the landmark ${\textbf{\textit l}}_i={[  x^i_k \ y^i_k]}^T$.}
    \label{fig:fig_1}
\end{figure}

Fig. \ref{fig:fig_1} shows the movement of a robot in a two-dimensional space and in Cartesian navigation coordinates. $x_k$ and $y_k$ are the robot coordinates and $\theta_k$ is the angle between the longitudinal line of the robot and the horizontal axis. We assume the robot observes $I$ landmarks at time $k$ that $x^i_k$ and $y^i_k$ are the coordinates of $i$th landmark. What the robot senses from the $i$th landmark is its distance to the robot $r^i_k$ and its viewing angle relative to the longitudinal line of the robot $\varphi^i_k$. A simple and general motion model for a steerable robot is presented based on instantaneous speed $v_k$ and steering angle $G_k$ of a robot as \cite{Bailey2004, Dissanayake2001}:

\begin{equation} \label{EQ_1}
\left[ \begin{array}{c}
x_{k+1} \\
y_{k+1} \\
\theta_{k+1} \end{array}
\right]=\left[ \begin{array}{c}
x_{k}+{v }_k \ d_t \ \cos(\theta_{k}+G_k) \\
{y_{k}+{v }_k \ d_t \ \sin(\theta_{k}+G_k)} \\
{\theta_{k}+{v}_k \ d_t \ \sin (G_k)\ }/D \end{array}
\right]+{{\mathbf w}}_{{ k}},
\end{equation}

\noindent where ${\mathbf x^r_k =[
\begin{array}{ccc}x_k & y_k & \theta_k
\end{array}]}^T$ is the robot state at time $k$ in a two-dimensional environment, and $\mathbf{w}_k$ is a vector of uncorrelated process noise with zero mean. $d_t$ is sampling period during time and $D$ is robot wheelbase. The robot state equation is nonlinear due to the presence of trigonometric functions of $\theta_{k}$ and $G_k$. So to use the Kalman filter, an approximation is needed to linearize the motion model.

In the entire SLAM implementation, the positions of landmarks observed so far must be added to the state vector; hence the state vector is extended as follows:
\begin{equation} \label{EQ_2}
\mathbf x_k ={[\  x_k   \ \  y_k  \ \ \theta_k \ \  x^1_k  \ \
    y^1_k \   \cdots \   x^n_k \ \  y^n_k \ ]}^T,
\end{equation}
\noindent where $n$ is the number of landmarks observed ever.

The observation from the $i$-th landmark with range $r^i_k$ and bearing $\varphi^i_k$ in the $k$-th step can be defined as:
\begin{equation} \label{EQ_3}
{{\mathbf z}}^i_k=\left[ \begin{array}{c}{\bar{r}}^i_k \\ \bar{\varphi}^i_k \end{array}\right]=\left[ \begin{array}{c}
\sqrt{{\left({x^i_k}-x_k\right)}^2+{\left({y^i_k}-y_k\right)}^2} \\
{{\arctan } \frac{\left({y^i_k}-y_k\right)}{\left({x^i_k}-x_k\right)}\ }-\ \theta_k \end{array}
\right]+{\mathbf n}^i_k,\hspace{0.1cm}   i \in \mathcal{I}_k,
\end{equation}
\noindent where $\mathcal{I}_k=\{i_1, i_2, ... , i_I \}$ denotes the numbers of landmarks under the observation of robot at time $k$, and ${\bar{r}}^i_k$ and $\bar{\varphi}^i_k$ are the measured/noisy values of ${{r}}^i_k$ and ${\varphi}^i_k$ with noise sensors measurement $\mathbf n^i_k$. It is clear that the observation model is also nonlinear with respect to state vector \eqref{EQ_2} and must be linearized to realize the Kalman filter. Note that the approximation caused by the linearization of the observation model is the main cause of EKF-SLAM divergence.

If we can model the motion and observations as a linear model, we have taken a big step towards reducing SLAM problems. For this purpose, we assume that the robot is equipped with a sensor that measures/estimates the robot angle $\theta_k$ in each step. This purpose is easily achieved by a compass. Meanwhile, we will show that the sensitivity of the proposed algorithm to the accuracy of angle measurement is very low.

\paragraph*{Linear model providing for motion}
First, according to \eqref{EQ_1}, the vector space of the robot is reduced to $x_k$ and $y_k$ as follows:
\begin{equation} \label{EQ_4}
\left[ \begin{array}{c}x_{k+1} \\ y_{k+1} \end{array}
\right]=\left[ \begin{array}{c}
x_{_{k}}+({\overline{v}}_{_{k}}+n^v_{k})\ d_t \ \cos(\bar{\theta}_{k}+n^{\theta }_k+{\overline{G}}_{k}+n^G_{k}) \\
y_{_{k}}+({\overline{v}}_{_{k}}+n^v_{k})\ d_t \ \sin(\bar{\theta}_{k}+n^{\theta }_k+{\overline{G}}_{k}+n^G_{k}) \end{array}\right],
\end{equation}
\noindent where $\overline{v}_k$, $\overline{G}_k$ and $\overline{\theta }_k$ are the measured/noisy values of $v_k$, $G_k$ and $\theta_k$. The corresponding measurement errors $n^v_k$, $n^G_k$ and $n^{\theta }_k$ are assumed to be independent zero mean Gaussian white noises with variances $\sigma^2_{v}$, $\sigma^2_G$ and ${\sigma }^2_{\theta }$ respectively. The equation \eqref{EQ_4} is linear in terms of state variables but non-linear in noises. Defining $n^{\psi}_k=n^{\theta }_k+n^G_k$ and by linearizing \eqref{EQ_4} in terms of $n^{\psi}_k$ and $n^v_k$ around zero the following equations are obtained as:
\begin{equation} \label{EQ_10}
\left[ \begin{array}{c}x_{k+1} \\ y_{k+1} \end{array}\right]=\left[ \begin{array}{c}x_{k} \\
y_{k} \end{array}\right]+{{\mathbf u}}_k+\mathbf G_k {\mathbf w}^n_{k},
\end{equation}
\noindent where ${\mathbf w}^n_{k}=\left[ n^v_k \hspace{0.2cm} n^{\psi}_k \right]^T$ is the Gaussian noise vector resulted from measurement error of the values of $v_k$, $\theta_k $ and ${ G_k}$ with zero mean and covariance matrix {$\mathbf Q_k=\text{diag} \{ {\sigma }^2_{v},   {\sigma }^2_{{\rm G}}{\rm +}{\sigma }^2_{\theta } \}$}. $\mathbf{u}_k$ is a given vector resulted from known input and $\mathbf G_k$ is the Jacobian matrix due to linearization \eqref{EQ_4} in terms of noises as follows:
\begin{equation} \label{EQ_5}
\mathbf u_k=\left[ \begin{array}{c}
\cos(\bar{\theta}_{k}+{\overline{G}}_{k}) \\
\sin(\bar{\theta}_{k}+{\overline{G}}_{k}) \end{array}\right]{\overline{v}}_{_{k}}\ d_t,
\end{equation}
\begin{equation} \label{EQ_9}
\mathbf G_k=\left[ \begin{array}{cc}{\cos(\bar{\theta}_{k}+{\overline{G}}_{k}) } & {-{\overline{v}}_{_{k}} \sin(\bar{\theta}_{k}+{\overline{G}}_{k}) } \\
{\sin(\bar{\theta}_{k}+{\overline{G}}_{k}) } & { {\overline{v}}_{_{k}} \cos(\bar{\theta}_{k}+{\overline{G}}_{k})  } \end{array}\right]d_t.
\end{equation}
By removing $\theta_k$ from state vector \eqref{EQ_2}, the new state vector will be as:
\begin{equation} \label{EQ_11}
\mathbf x_k ={[\  x_k   \ \  y_k  \ \  x^1_k  \ \
    y^1_k \   \cdots \   x^n_k \ \  y^n_k \ ]}^T,
\end{equation}
and since all of the landmarks are considered fixed, the motion equation of SLAM can be written as:
\begin{equation} \label{EQ_13}
{{\mathbf x}}_{k+1}={{\mathbf x}}_k+{\left[ \begin{array}{c}
    {{\mathbf u}}_k \\ \mathbf 0_{2n\times1} \end{array}
    \right]}+{\left[ \begin{array}{c}\mathbf G_k {\mathbf w}^n_{k} \\ \mathbf 0_{2n\times1} \end{array}\right]}.
\end{equation}
\noindent As be seen, the final motion model \eqref{EQ_13} is linear in terms of state variables and measurement noises of sensors.

\paragraph*{Linear model providing for observations}
The initial observation model is presented by \eqref{EQ_3}. We assume that the robot observes $I$ landmarks in the \textit{k}-th step whose numbers are in $\mathcal{I}_k$ set. For the \textit{i}-th landmark, the data is ${\overline{r}}^i_k$ and ${\overline{\varphi }}^i_k$, $i\in\mathcal{I}_k$. As the robot angle ${\overline{\theta}}_k$ is measured in the proposed method, the observation model can be easily linearized by the following simple transformation:
\begin{equation} \label{EQ_14}
{{\boldsymbol{\zeta}}}^i_k=\mathbf{h}({{\mathbf z}}^i_k)=\left[ \begin{array}{c}
{\overline{r}}^i_k{\cos  ({\overline{\theta }}_k+{\overline{\varphi }}^i_k)\ } \\
{\overline{r}}^i_k{\sin  ({\overline{\theta }}_k+{\overline{\varphi }}^i_k)\ } \end{array}
\right]    \ \   ,  \ \ \ \  i\in{\mathcal{I} }_k.
\end{equation}
Since the measured data ${\overline{\theta }}_k$, ${\overline{r}}^i_k$ and ${\overline{\varphi }}^i_k$ have uncertainties due to measurement errors, the equation \eqref{EQ_14} can be written as:
\begin{equation} \label{EQ_15}
{{\boldsymbol{\zeta}}}^i_k=\left[ \begin{array}{c}
(r^i_k+n^{ir}_k) \ {\cos ({\theta }_k+{\varphi }^i_k+n^{\theta }_k+n^{i\varphi }_k)\ } \\
(r^i_k+n^{ir}_k) \ {\sin({\theta }_k+{\varphi }^i_k+n^{\theta }_k+n^{i\varphi }_k)\ } \end{array}\right],
\end{equation}
\noindent where $n^{\theta }_k$, $n^{ir}_k$ and $n^{i\varphi}_k$ are independent white Gaussian noises with zero mean and variances $\sigma^2_\theta$, $\sigma^2_r$ and $\sigma^2_\varphi$ resulted from robot angle, range and bearing measurements respectively. Defining $n^{i\omega}_k=n^{\theta }_k+n^{i\varphi }_k$ and by linearizing \eqref{EQ_15} in terms of $n^r_k$ and $n^{i\omega}_k$ around zero, the following equations are resulted as:

\begin{equation} \label{EQ_18}
{{\boldsymbol{\zeta}}}^i_k\cong \left[ \begin{array}{c}
r^i_k{\cos  \left({\theta }_k+{\varphi }^i_k\right)\ } \\
r^i_k{\rm sin}\left({\theta }_k+{\varphi }^i_k\right) \end{array}
\right]+\mathbf S^i_k {\boldsymbol\vartheta}^i_k,
\end{equation}
\noindent where ${\boldsymbol\vartheta}^i_k=\left[ n^{ir}_k \hspace{0.2cm} n^{i\omega}_k \right]^T$ is the Gaussian noise vector resulted from measurement error of the values of ${r^i_k}$, $\theta_k $ and $\varphi^i_k$ with zero mean and covariance matrix {$\mathbf R^i_k= \rm{diag} \{ {\sigma }^2_{r} , {\sigma }^2_{{\rm \varphi}}{\rm +}{\sigma }^2_{\theta } \}$}. $\mathbf S^i_k$ is the Jacobian matrix due to linearization \eqref{EQ_15} in terms of noises as follows:
\begin{equation} \label{EQ_17}
\mathbf S^i_k=\left[ \begin{array}{cc}
{\cos  ({\overline{\theta }}_k+{\overline{\varphi }}^i_k)\ } & -{\overline{r}}^i_k{\sin  ({\overline{\theta }}_k+{\overline{\varphi }}^i_k)\ } \\
{\sin  ({\overline{\theta }}_k+{\overline{\varphi }}^i_k)\ } & {\overline{r}}^i_k{\cos  ({\overline{\theta }}_k+{\overline{\varphi }}^i_k)\ } \end{array}
\right].
\end{equation}
\noindent According to Fig. \ref{fig:fig_1}, the following relations hold between the state variables $x_k$, $y_k$, $x^i_k$, $y^i_k$ and observed variables $r^i_k$, ${\varphi }^i_k$:
\begin{equation} \label{EQ_19}
\begin{split}
x^i_k-x_k=r^i_k \cos({\theta }_k+{\varphi }^i_k), \\
y^i_k-y_k=r^i_k \sin({\theta }_k+{\varphi }^i_k).
\end{split}
\end{equation}
\noindent Substituting \eqref{EQ_19} to \eqref{EQ_18}, the following linear equations can be resulted for the observations vector ${{\boldsymbol{\zeta}}}^i_k$:
\begin{equation} \label{EQ_20}
{{\boldsymbol{\zeta}}}^i_k=\left[ \begin{array}{c}x^i_k-x_k\\
y^i_k-y_k \end{array}\right]+\mathbf S^i_k{ \boldsymbol\vartheta}^i_k={{\mathbf H}}_k^i{{\mathbf x}}_k+\mathbf S^i_k \boldsymbol {\vartheta}^i_k \ \  , \ \   i\in{\mathcal{I}}_k,
\end{equation}
\noindent where observation matrix ${{\mathbf H}}^i$ is as below:

\begin{equation} \label{EQ_22}
\mathbf H^i_k =\left[\ -1   \ \  \delta_{i-1}  \ \ \cdots  \   \delta_{i-n}\ \right]\otimes \mathbf I_{2} \hspace{.2cm} ,  i\in{\mathcal{I}}_k,
\end{equation}
and $\delta_j$ denotes the Kronecker delta function and $\otimes$ is Kronecker product.

\noindent As be seen, the final observation model \eqref{EQ_20} is linear in terms of state variables and measurement noises similar to the motion model \eqref{EQ_10}. Therefore, the Kalman filter can be directly applied to this model.

Based on linear model \eqref{EQ_13} and \eqref{EQ_20}, the filtering algorithm is summarized in Algorithm~\ref{Alg1}. There, $\hat{\mathbf x}_{k_1|k_2}$ denotes the estimation of ${\mathbf x}_{k_1}$ from the data up to the time $k_2$ and ${\mathbf \Sigma}_{k_1|k_2}$ is its covariance matrix. Note that since the data from different landmarks are independent, we use the sequential processing KF for updating the state vector. This algorithm increases the filtering speed significantly.

\begin{algorithm}[h]
    \begin{algorithmic}[1]
        \renewcommand{\algorithmicrequire}{\textbf{{Initial values:}}}
        \Require {$\left[\begin{array}{c}
        \hat{\mathbf x}^0  \\ {\mathbf \Sigma}^0 \end{array}\right]=\left[\begin{array}{c}
        \hat{\mathbf x}_{k|k-1}  \\  {\mathbf \Sigma}_{k|k-1} \end{array}\right]$}
        \State {\textbf{Data update step} by \eqref{EQ_20}:
            \\ for $i$ from $1$ to $N_I=|{\mathcal{I}}|$ :\\
            $\hat{\mathbf x}^i =\hat{\mathbf x}^{i-1} +{{\mathbf L}}^i({{\boldsymbol{\zeta}}}^i_k-{{\mathbf H}}_k^i\hat{\mathbf x}^{i-1} )$\\
            ${{\mathbf L}}^i={\mathbf \Sigma}^{i-1}{{{\mathbf H}}_k^i}^T ({{{\mathbf H}}_k^i}{\mathbf \Sigma}^{i-1}{{{\mathbf H}}_k^i}^T+{{{\mathbf S}}_k^i}{{{\mathbf R}}_k^i}{{{\mathbf S}}_k^i}^T)^{-1}$\\
            ${\mathbf \Sigma}^i={\mathbf \Sigma}^{i-1}-{{\mathbf L}}^i{{{\mathbf H}}_k^i}{\mathbf \Sigma}^{i-1}$\\
            End for \\
            $\left[\begin{array}{c}
            \hat{\mathbf x}_{k|k}  \\  {\mathbf \Sigma}_{k|k} \end{array}\right]=\left[\begin{array}{c}
            \hat{\mathbf x}^{N_I}  \\ {\mathbf \Sigma}^{N_I} \end{array}\right]$}
        \State {\textbf{Prediction Step} by \eqref{EQ_13}
            \\ $\hat{\mathbf x}_{k+1|k} =\hat{\mathbf x}_{k|k} +[{{\mathbf u}}^T_k \hspace{0.1cm}\mathbf 0_{1\times2n} ]^T$\\
            ${\mathbf \Sigma}_{k+1|k}={\mathbf \Sigma}_{k|k}+\text{blkdiag}\{ \mathbf G_k \mathbf Q_k  \mathbf G_k^T,  \mathbf 0_{2n\times2n}  \}$.}
    \end{algorithmic}
    \caption{{Data Update and Prediction Steps.}}\label{Alg1}
\end{algorithm}

\noindent In section 3, it will be shown that the proposed method is significantly superior to state-of-the-art methods, especially EKF-SLAM. We see even if the steering sensor \textit{G} is not used in the robot motion equation, the proposed method still works better than the standard EKF-SLAM.

It is very hard to find mathematical solutions to prove the convergence of the EKF-SLAM method which has nonlinear models. So simulation is the only way to evaluate all of such types of methods. However, in the proposed method, since the motion and observation models are linear, the study of its convergence is possible. The next subsection is dedicated to this discussion.

\subsection{Convergence Study of the Proposed SLAM Method}

\noindent Dissanayake et al. investigated the convergence of SLAM based on Kalman filter for the first time \cite{Dissanayake2001}. They proved that the absolute positioning error of landmarks is not grown with each observation and finally tends towards a lower bound value. In this proof, the motion and observation models have been considered linear. However, in most introduced practical SLAM techniques, the motion and observation models of robot are nonlinear. In most of these studies, EKF-based SLAMs have been considered. In \cite{Julier2001}, the divergence of EKF SLAM has been reported, even when the robot is stationary. Many researchers have tried to reduce the SLAM divergence \cite{Castellanos2007, Huang2010, Barrau2015}, however, no convergence guarantee has yet been reported. Since the motion and observation models of our proposed method are linear, the proof of convergence in \cite{Dissanayake2001} is also valid for our method.

In the next section, the performance of the proposed method is studied. It is shown that even after thousands of tests in different environments and under different conditions, there would be no sign of divergence of the proposed method.

\section{Experiments}\label{sec3}

We evaluate the performance of our method and compare it with the standard EKF, UnFS-SLAM, and ICKF methods. The volume of calculations and the accuracy of UnFS-SLAM depends on the number of particles in implementation. In order that the volume of calculation of the UnFS-SLAM is around the volume of other methods, five particles are allocated for it. In these situations, the computation volume of the UnFS-SLAM is nearly three times the LMKF and EKF methods. Our experiments have been carried out in two different environmental conditions. \noindent In the first tests, an artificial environment with virtual landmarks is created by modified Baily simulation software \cite{Bailey2004} and Kim software\footnote{https://svn.openslam.org/data/svn/ufastslam}. In these tests, the effects of changes in simulation parameters are examined. In the second test, which is performed on real data set on Sydney Victoria Park dataset\footnote{The data set is available at http://www-personal.acfr.usyd.edu.au/nebot/victoria\_park.htm.}, there is no possibility of changing the experimental conditions and parameters presented in the dataset are used.

To test on real data, outdoor datasets are preferred to indoor datasets because in indoor environments the robot speed is usually low and the robot commonly moves over a flat and smooth surface, so the motion sensors' uncertainties are low (vibrations on the move have minor effects on sensors errors). These reasons cause the divergence of EKF and UnFS SLAM methods is postponed.

\subsection{The Simulation Results}

\noindent An artificial environment with $200\times200\hspace{.05cm} m^2$ area is used for the first test. The robot's initial position is $[0, 0]^T$. Other flexible parameters which can be changed in simulation calculation are listed in Table.\ref{TB_1}. In this simulation, each test with its own fixed parameters is performed at least 50 times and the final results are averaged to eliminate the effect of different noise sequences.

\begin{table}[h]
    \centering
    \renewcommand{\arraystretch}{1.1}
    \caption{Simulation parameters setting \cite{Bailey2004}.}
    \label{TB_1}
    \begin{tabular}{ p{3.5in} p{.5in} } \hline
        Wheelbase of Robot \textit{D} & $1m$ \\ \hline
        Control Frequency ${1}/{d_t}$ & $10\ Hz$ \\ \hline
        Observation Period ${\Delta}t_{{o}}$ & $2\ Sec$ \\ \hline
        Maximum Observation Sensor Range & $25m$ \\ \hline
        Maximum Steering Angle & ${30}^{^\circ }$ \\ \hline
        Standard deviation of Steering Angle $\sigma_{_{G}}$ & $1^{^\circ }$ \\ \hline
        Standard deviation of robot angle Sensor $\sigma_{_{\theta}}$ & $1^{^\circ }$ \\ \hline
        Standard deviation of Observation Sensors $\sigma_{_{r}}$, ${\sigma }_{{\rm \varphi}}$ & $0.1 \ m , \ 1^o$ \\ \hline
        Robot speed $v$ & $1m/s$ \\ \hline
        Standard deviation of Robot Speed $\sigma_{v}$ & $0.1m/s$ \\ \hline
    \end{tabular}
\end{table}

As the main goal is the comparison between the efficiency and abilities of the proposed LMKF SLAM with EKF, UnFS, and ICKF SLAMs, the root mean square error (RMSE) of robot and landmarks positioning is used as a comparison criterion. The behavior of SLAM methods in closed-loop and open-loop paths is different because in paths with several closed loops, it can compensate their previous mistakes. So a better comparison of functionality and convergence of SLAM methods would be achieved in paths with no closed loops.

\noindent To make a comparison, two different paths are considered; the first path with several closed loops with $1350\hspace{0.05cm}m$ length and 139 landmarks and the second path without closed loops, with $577\hspace{0.05cm}m$ length and 63 landmarks.

\begin{figure}[h]
    \centering
    \begin{subfigure}{.49\textwidth}
        \centering
        \includegraphics[width=1.\linewidth,height=2.2in]{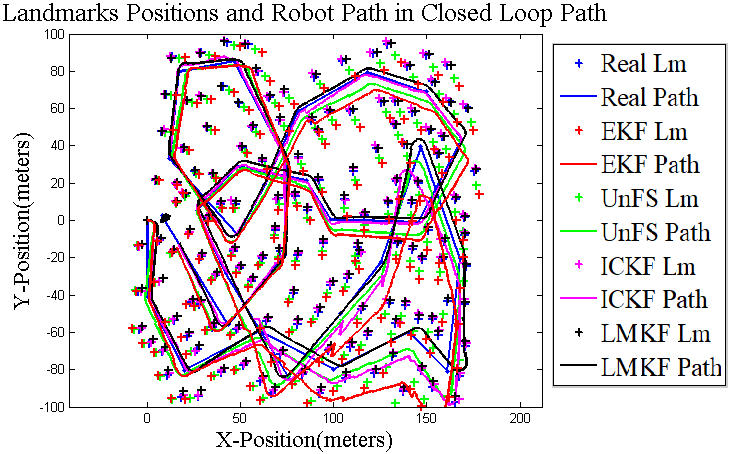}
        \caption{}
        \label{fig_2a}
    \end{subfigure}
    \begin{subfigure}{.49\textwidth}
        \centering
        \includegraphics[width=1.\linewidth,height=2.2in]{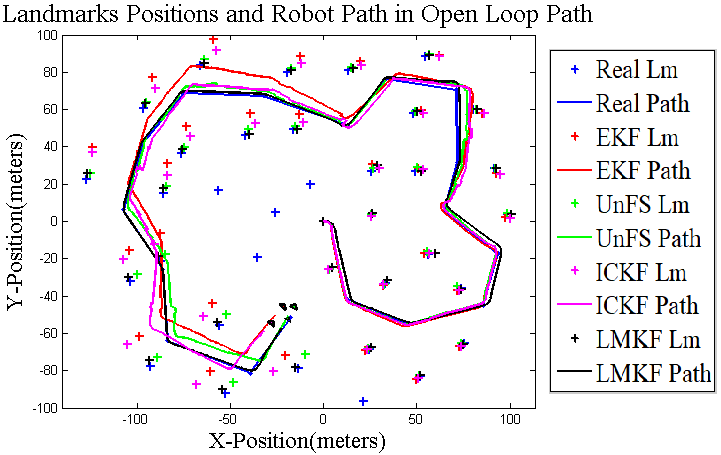}
        \caption{}
    \end{subfigure}
    \caption{The simulated test environment with specified positions for landmarks and the robot path (blue) by the LMKF, EKF, UnFS, and ICKF SLAMs. The landmarks (Lm) are marked with +. a) a closed-loop path with $1350\hspace{0.05cm}m$ length and 139 landmarks, b) an open-loop path with $577\hspace{0.05cm}m$ length and 63 landmarks.}
    \label{fig:fig_2}
\end{figure}

Fig. \ref{fig:fig_2} shows these two defined paths and landmarks along with the maps estimated by the mentioned methods. According to Fig. \ref{fig:fig_2}, it is obvious that in all methods, after each closed-loop, the robot reduces the estimation errors and could get closer to the real path. However, it is obvious that the LMKF significantly outperforms other methods. The ICKF, UnFS, and EKF are the next ranks.

Our investigation shows that the steering angle sensor \textit{G} does not have much effect on the performance of LMKF SLAM; hence we ignore $G$ in the LMKF equations. We study the effects of different parameters such as $v$ and $\ \sigma_v$ on RMSE of robot and landmarks positions based on simultaneous changes of ${\sigma }_{_G}$, ${\sigma }_{_{\theta}}\ $in each method. ${\sigma }_{_G}$ affects only on the EKF and UnFS and ${\sigma }_{_{\theta}}$ affects on LMKF.

\begin{figure}[h]
    \centering
    \begin{subfigure}{.49\textwidth}
        \centering
        \includegraphics[width=1.\linewidth,height=2.2in]{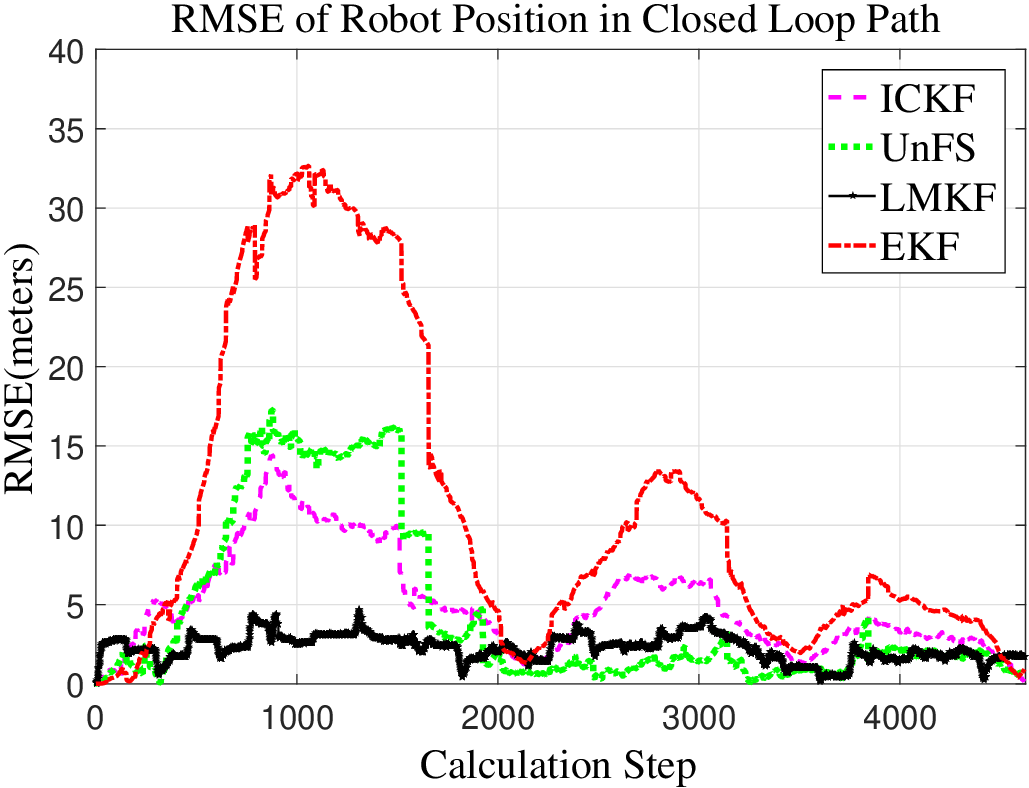}
        \caption{}
        \label{fig_3a}
    \end{subfigure}
    \begin{subfigure}{.49\textwidth}
        \centering
        \includegraphics[width=1.\linewidth,height=2.2in]{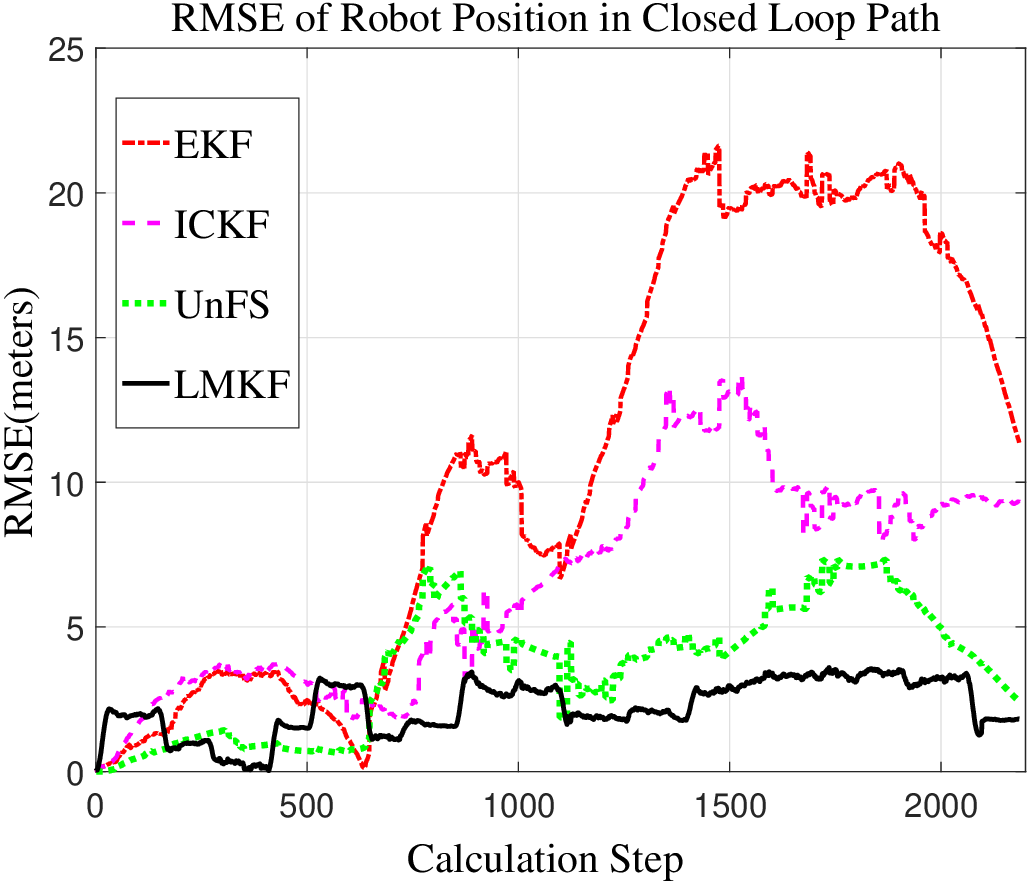}
        \caption{}
    \end{subfigure}
    \caption{The RMSE of robot position for the EKF, UnFS, ICKF, and LMKF methods over time travel of robot with conditions: $v= 2 m/s$, ${\sigma }_v=0.2m/s$, ${\sigma }_{_G}\ $= $2^o$, ${\sigma }_{_{{\theta }}}=2^o$ a) closed-loop b) open-loop paths.} \label{fig:fig_3}
\end{figure}

Fig. \ref{fig:fig_3} shows the RMSE of robot position for EKF, UnFS, ICKF, and LMKF over the traveling time of two defined paths in Fig. \ref{fig:fig_2} and based on setting parameters in Table. \ref{TB_1}. It is seen that the performance of LMKF significantly outperforms the other methods. Its performance is with few changes even in an open-loop path. Although in the closed-loop path, the error of other methods is decreased when previously observed landmarks are revisited; but in a path without loops, their error reduction is local and insignificant. The results show that the performance of LMKF is so good that the existence of closed loops in paths has low effects on the accuracy of robot and landmarks positioning.

\begin{figure}[h]
    \centering
    \begin{subfigure}{.49\textwidth}
        \centering
        \includegraphics[width=1\linewidth,height=2.2in]{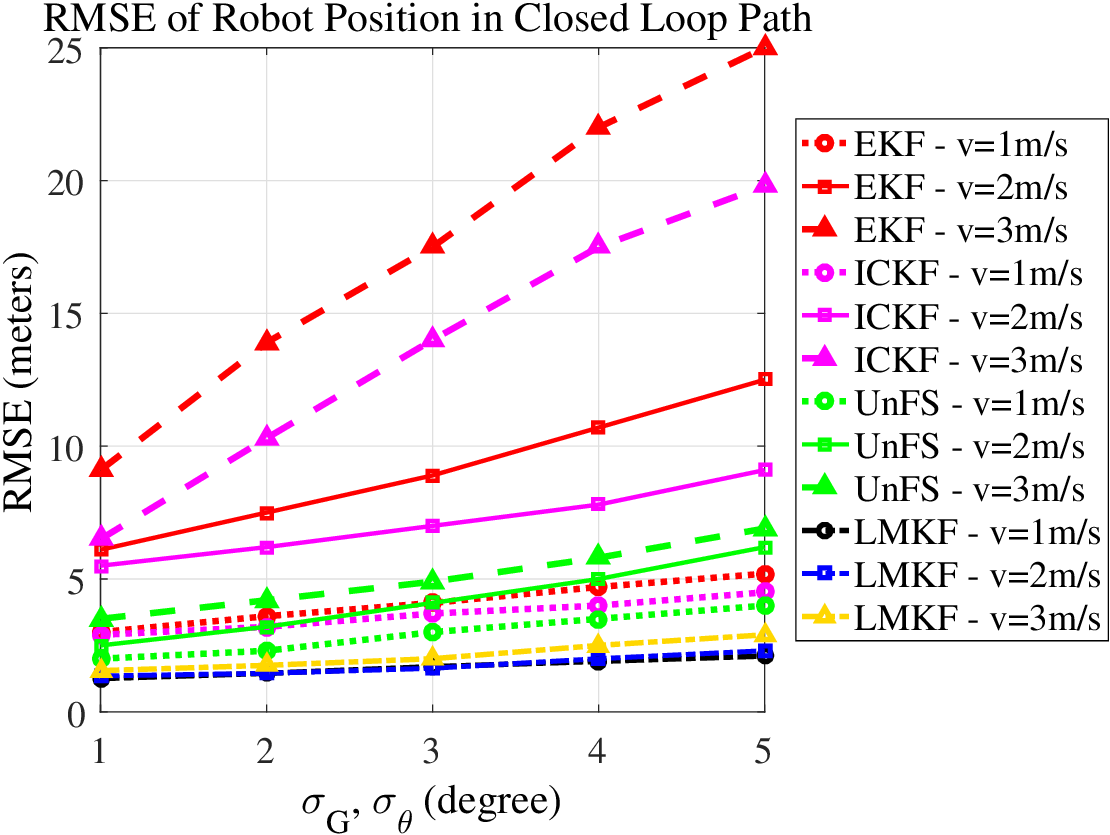}
        \caption{}
        \label{fig_4a}
    \end{subfigure}
    \begin{subfigure}{.49\textwidth}
        \centering
        \includegraphics[width=1\linewidth,height=2.2in]{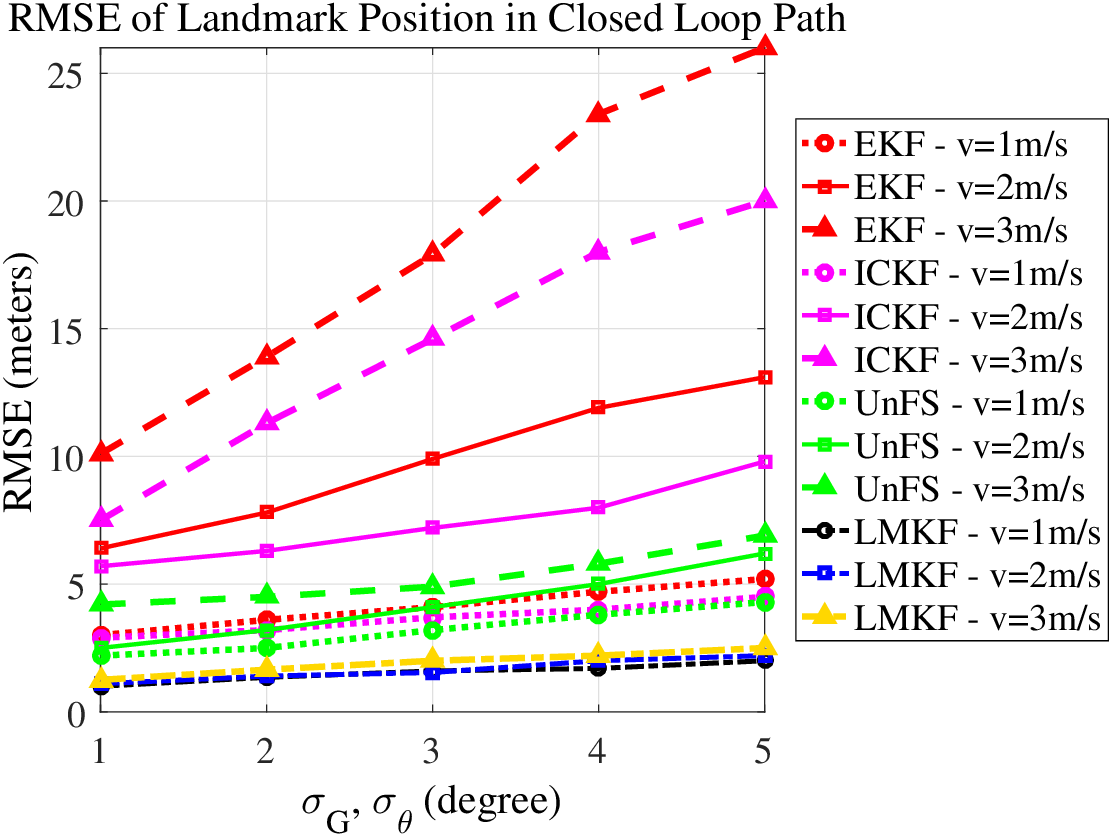}
        \caption{}
    \end{subfigure}
    \caption{a) RMSE of robot position and b) RMSE of landmarks positions, for EKF, UnFS, ICKF, and LMKF methods in the closed-loop path with different $v=1,\ 2,\ 3\ m/s$ and ${\ \sigma }_v=0.1 m/s$ versus ${\sigma }_{_G}$ and $\sigma_{_{{\theta }}}$.}
    \label{fig:fig_4}
\end{figure}

If RMSE values of Fig. \ref{fig:fig_3} are averaged over the traveling time of the robot, and the test process is performed each time with different parameters listed in Table. \ref{TB_1}, the average of RMSE can be calculated for tests with different conditions. Fig. \ref{fig:fig_4} shows the RMSE of the robot and landmarks positions for the LMKF, UnFS, ICKF, and EKF methods in the closed-loop path. This test is carried out based on changes in $v$, ${\sigma }_{_G}$ and ${\sigma }_{_{{\theta }}}$. In this test, ${\ \sigma }_v$ is considered equal to $0.1 m/s$. For each set of considered test parameters, the test process has been performed 50 times and finally, the average of RMSE has been calculated. Fig. \ref{fig:fig_4} clearly shows that the RMSE in EKF and ICKF grows rapidly with increasing ${\sigma }_{_G}$ and $v$ and with less growth in UnFS. However, in LMKF, the RMSE changes are insignificant when $v$ is increased, such that RMSE values at different speeds can hardly be distinguished from each other. Also, when the ${\sigma }_{_{{\theta }}}$ is increased, the growth of RMSE is little and negligible. This figure shows that the proposed method unlike the other methods is robust to the changes of the robot speed and angle uncertainty.

\begin{figure}[h]
    \centering
    \begin{subfigure}{.49\textwidth}
        \centering
        \includegraphics[width=1\linewidth,height=2.2in]{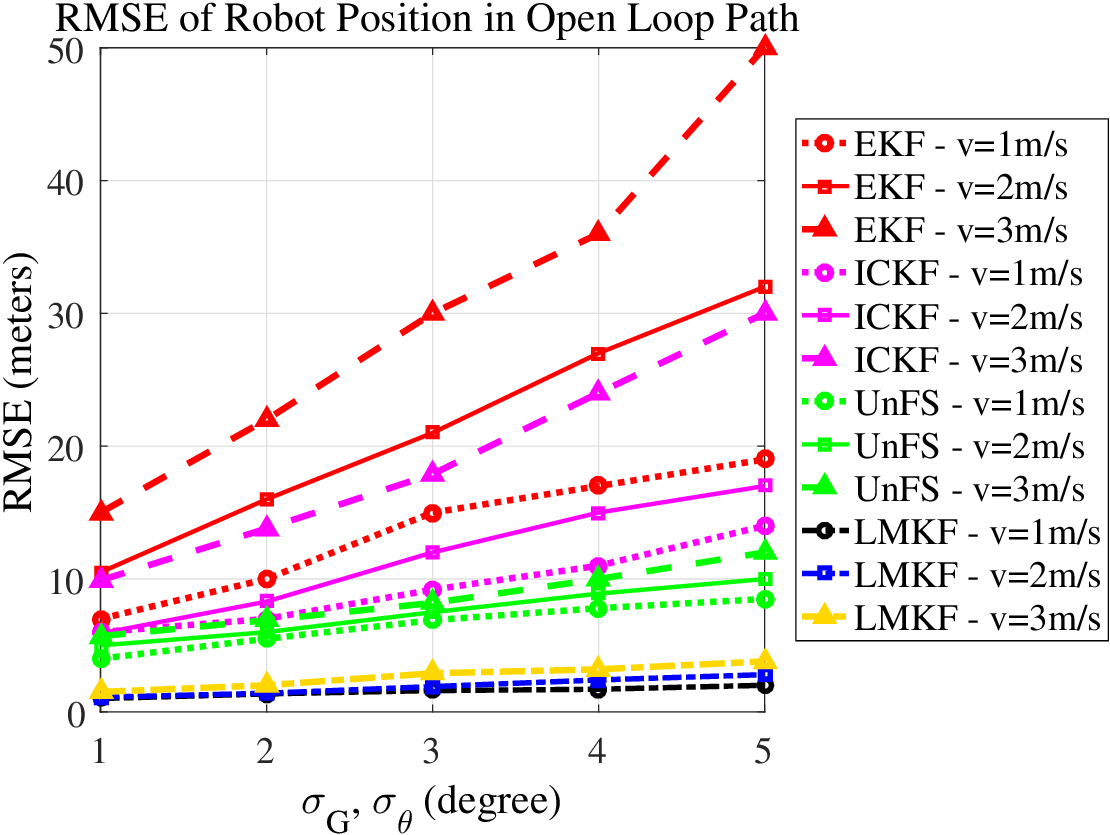}
        \caption{}
        \label{fig_5a}
    \end{subfigure}
    \begin{subfigure}{.49\textwidth}
        \centering
        \includegraphics[width=1.1\linewidth,height=2.2in]{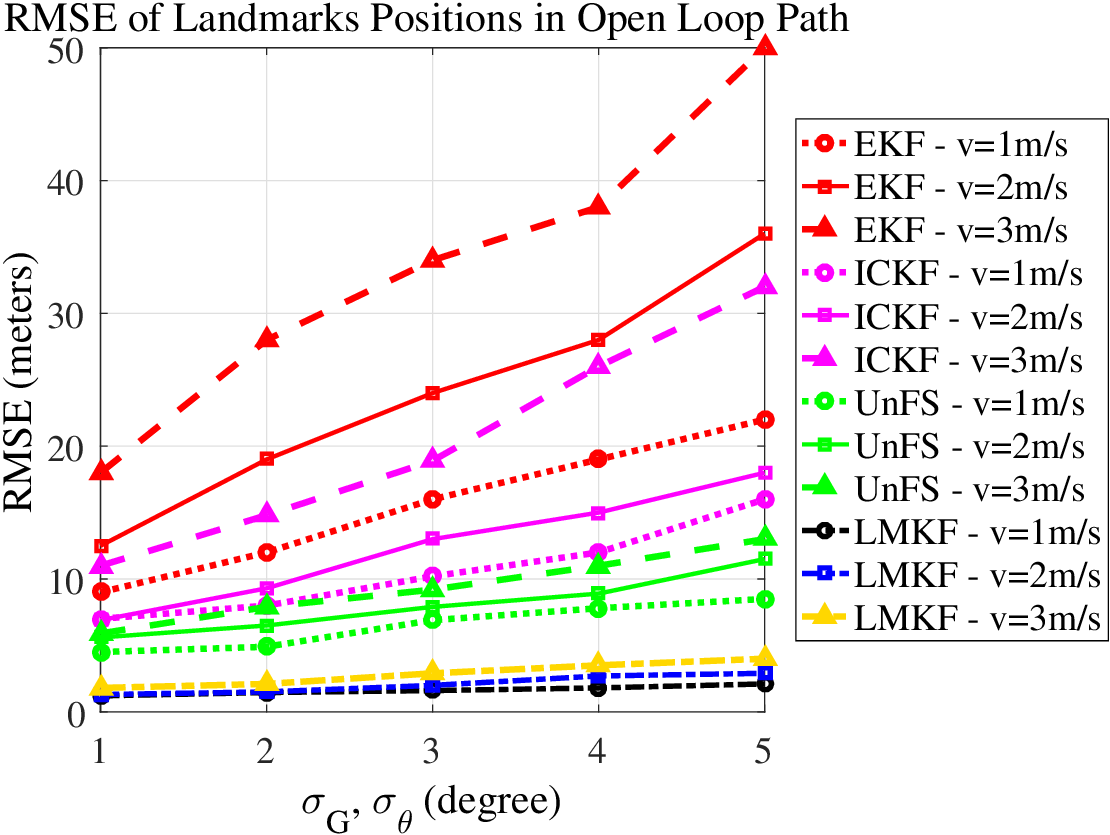}
        \caption{}
    \end{subfigure}
    \caption{a) RMSE of robot position and b) RMSE of landmarks positions, for EKF, UnFS, ICKF, and LMKF in the open-loop path in different $v=1,\ 2,\ 3\ m/s$ and with ${\sigma }_v=0.1 m/s$ versus ${\sigma }_{_G}$ and $\sigma_{_{{\theta }}}$.}
    \label{fig:fig_5}
\end{figure}

Fig. \ref{fig:fig_5} shows the RMSE of the robot and landmarks positions for the mentioned methods in the open-loop path. The simulation conditions are the same as the previous test. It can be seen while the RMSE of LMKF has been changed a little in comparison to the previous test, the RMSE of other methods increases in comparison with the results of Fig.\ref{fig:fig_4}. The RMSE in EKF, ICKF, and UnFS in different speeds of the robot is 6 to 15, 5 to 10, and 3 to 4 times of the RMSE in LMKF, respectively. This means that the error of the proposed method does not highly depend on the closed loops existence in the path.

\begin{figure}[h]
    \centering
    \begin{subfigure}{.49\textwidth}
        \centering
        \includegraphics[width=1\linewidth,height=2.2in]{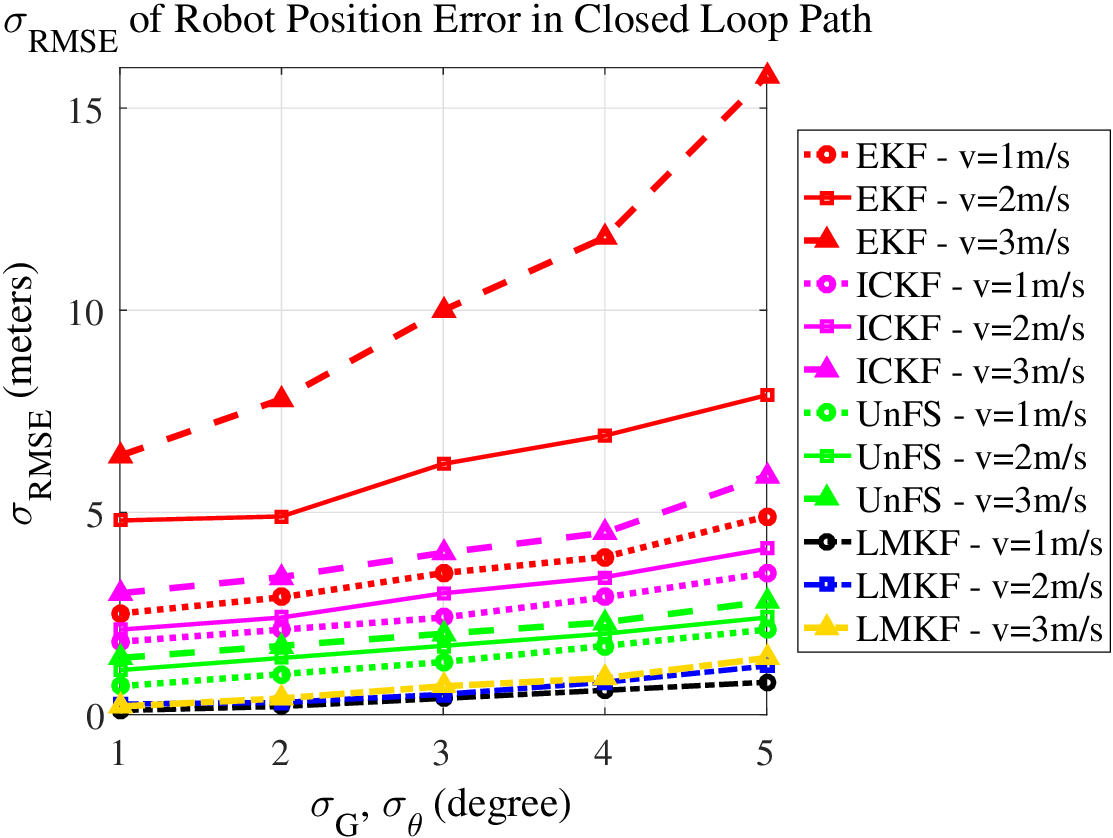}
        \caption{}
        \label{fig_6a}
    \end{subfigure}
    \begin{subfigure}{.49\textwidth}
        \centering
        \includegraphics[width=1\linewidth,height=2.2in]{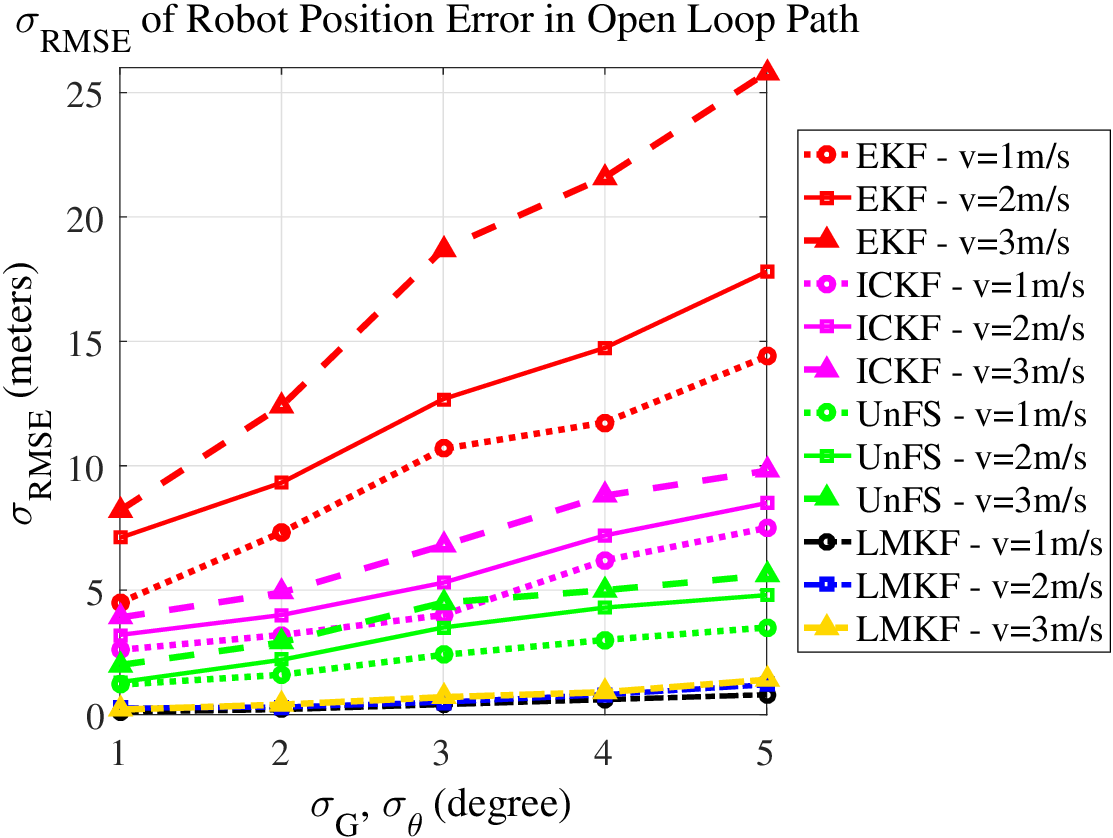}
        \caption{}
    \end{subfigure}
    \caption{${\sigma }_{_{ RMSE}}$ of robot position for EKF, UnFS, ICKF, and LMKF methods with different $v=1,\ 2,\ 3\ m/s$ and ${ \sigma }_v=0.1 m/s$ versus ${\sigma }_{_{_G}}$ and $\sigma_{_{{\theta }}}$ a) closed-loop path, b) open-loop path.}
    \label{fig:fig_6}
\end{figure}

The stability of a method to the different noise sequences is an important criterion which is shown as the error variance in different tests. Fig. \ref{fig:fig_6} shows the standard deviation of RMSE of the robot and landmarks positioning ${\sigma }_{_{RMSE}}$. The values have resulted after running the simulation for 50 times on closed-loop and open-loop paths. As be seen in Fig. \ref{fig:fig_6}, the values and incremental rate of ${\sigma }_{_{RMSE}}$ for the EKF are larger than that of other methods, both in the closed loop and open loop paths. However, the LMKF has significantly the lowest values and incremental rate of ${\sigma }_{_{RMSE}}$. This indicates that the LMKF method not only has a lower RMSE but also has more reliability, and it can have low variations based on the noise sequence changes with respect to the other methods.

\begin{figure}[h]
    \centering
    \begin{subfigure}{.49\textwidth}
        \includegraphics[width=1\linewidth,height=2in]{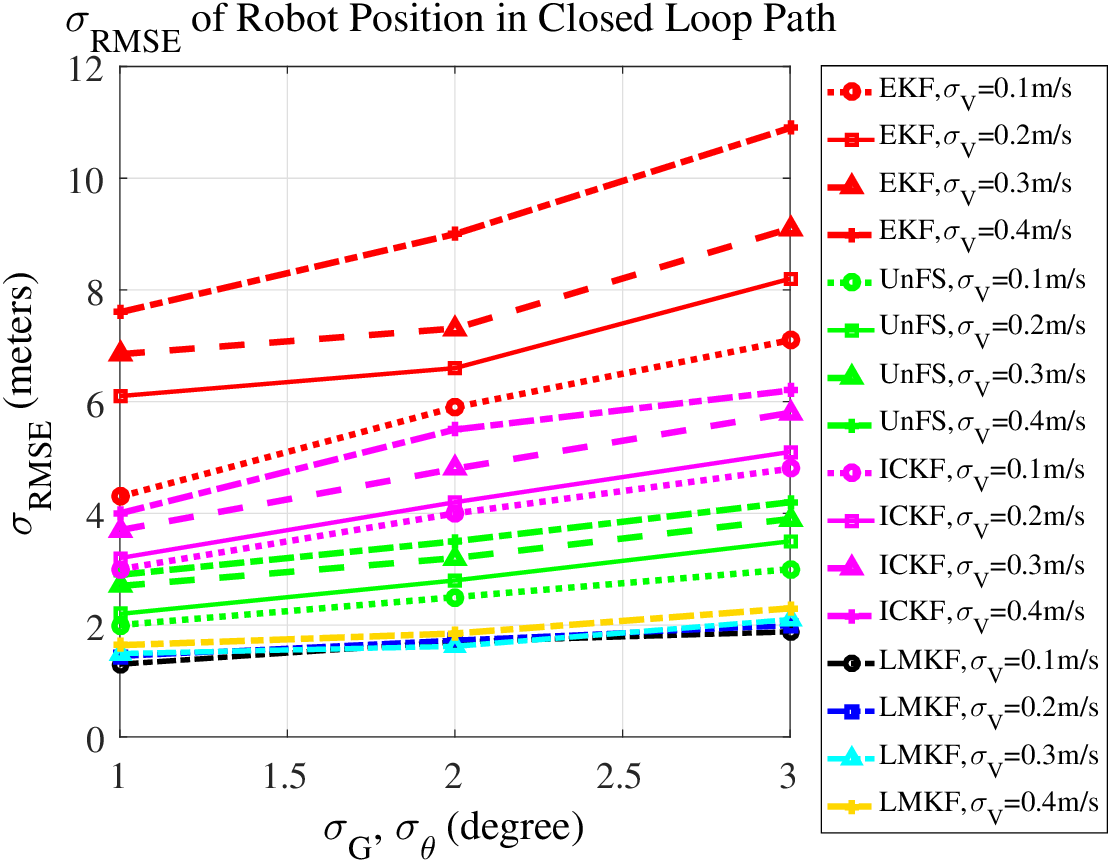}
        \caption{}
        \label{fig_7a}
    \end{subfigure}
    \begin{subfigure}{.49\textwidth}
        \includegraphics[width=1\linewidth,height=2in]{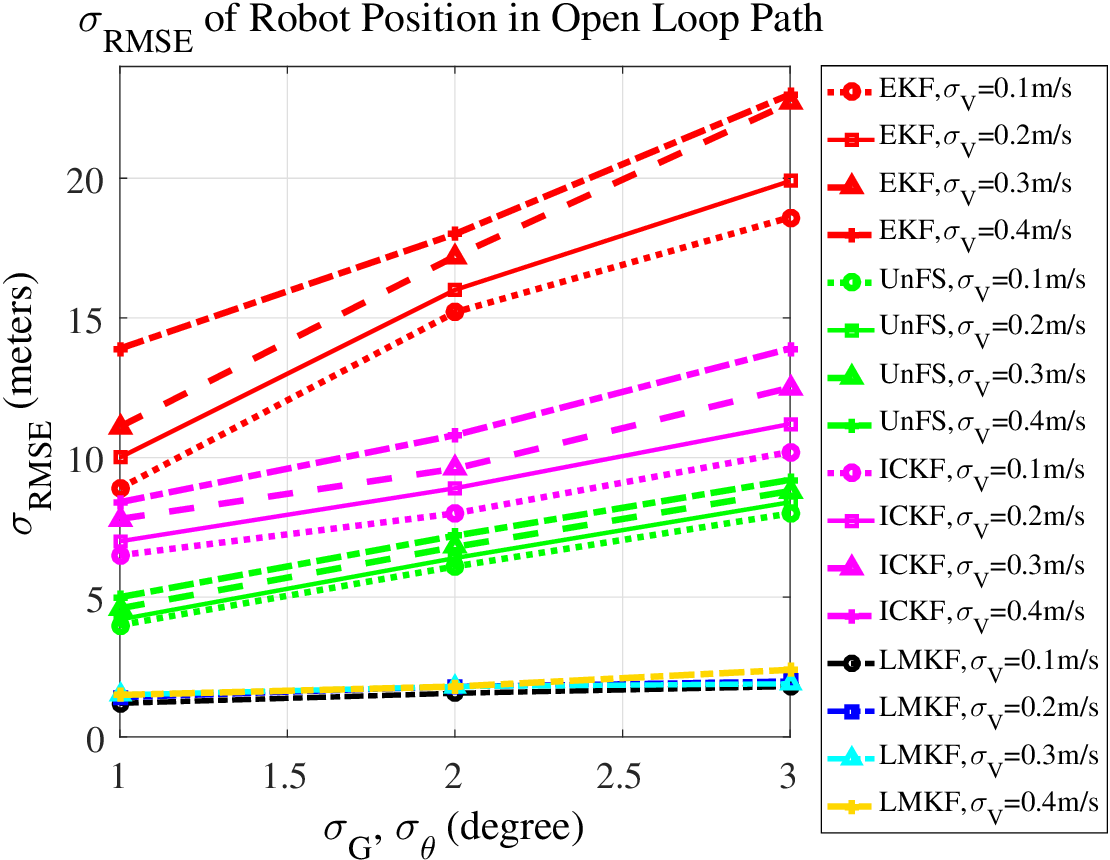}
        \caption{}
    \end{subfigure}
    \caption{${\sigma }_{_{RMSE}}$ of robot position for EKF, UnFS, ICKF, and LMKF with $v=\ 2m/s$ and different ${\sigma }_v=0.1,0.2,0.3,0.4\ m/s$ versus ${\sigma }_{_G}$ and $\sigma_{_{{\theta }}}$ a) closed-loop path, b) open-loop path.}
    \label{fig:fig_7}
\end{figure}

In previous tests, the uncertainty of robot speed ${\sigma }_v$ is considered $0.1 m/s$. In Fig. \ref{fig:fig_7}, we show the RMSE of robot position relative to ${\sigma }_{_G}$ and $\sigma_{_{{\theta }}}$, for different values of ${\sigma }_v$ in the mentioned SLAM methods on paths defined in Fig.\ref{fig:fig_2}. Figure \ref{fig:fig_7} shows that LMKF is robust against ${\sigma }_v$ changes and has insignificant variations, while ${\sigma }_{RMSE}$ of EKF, UnFS, and ICKF increases significantly with the growth values of ${\sigma }_v$.

\subsection{Experimental Results}

\noindent To provide the test process of the proposed method with real information, the Sydney Victoria Park dataset has been used. This dataset includes the measured information by a two-dimensional laser scanner, a GPS, and a sensor to measure the speed and bearing angle of the vehicle. In this data set, the real position of landmarks, which are trees is not known. Hence the only criterion to compare the performance of methods can be the comparison of the vehicle localization accuracy with GPS information in areas where GPS information is available.

\begin{figure}[h]
    \centering
    \includegraphics*[width=6.5cm, height=6.5cm]{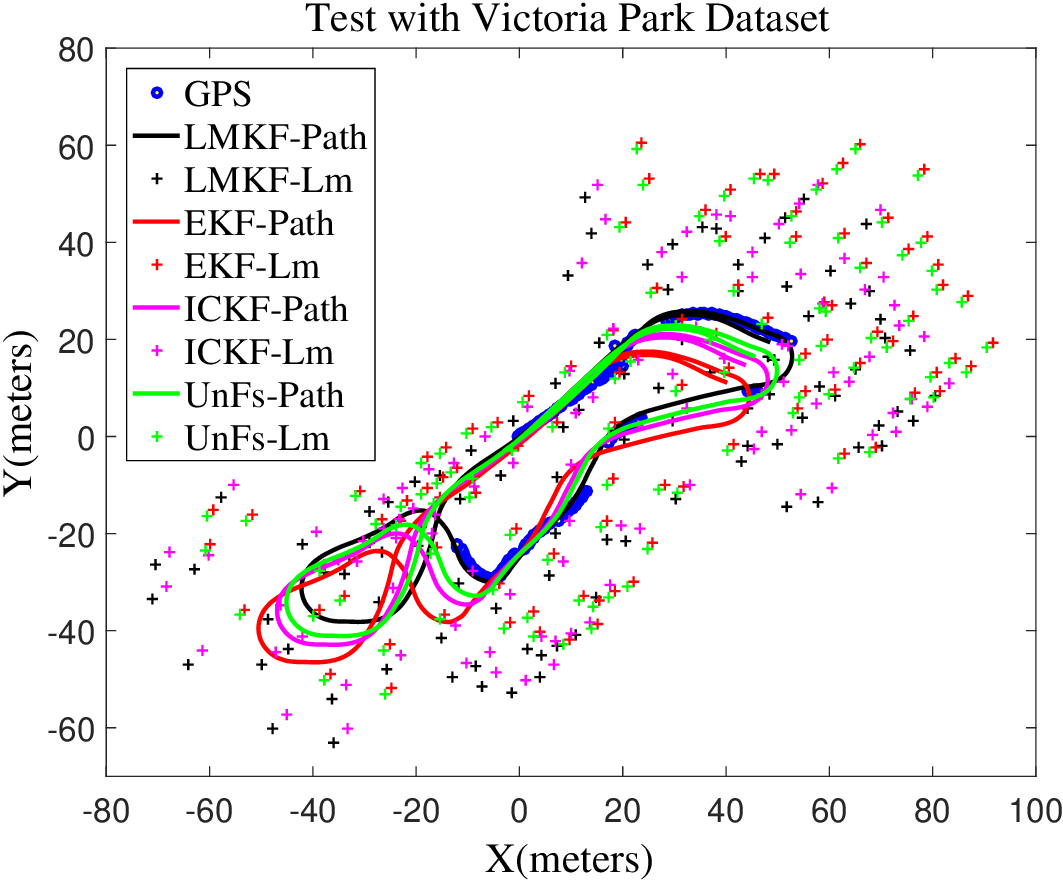}
    \caption{Landmarks and vehicle positions extracted in the simulation based on the Sydney Victoria Park dataset.} \label{fig:fig_10}
\end{figure}

The information from 620 primary observations of this dataset is used to this comparison. In Fig. \ref{fig:fig_10}, the test environment conditions which include about 100 landmarks and the vehicle path have been presented. In this figure, the places where GPS information is available have been specified too. It can be seen that this information is not available continuously throughout the path of the vehicle. In addition, sometimes a spot noise has occurred in determining the vehicle position.

To implement the LMKF method, a robot angle sensor measurement must be used however this information is not included in this dataset. To perform this test, GPS and the path of vehicle information are used to estimate the data of a robot angle sensor that does not exist. The lack of GPS information in some parts of the path and spot noise existence cause the errors of this estimation to become large. This low accuracy of estimation leads to the low precision of the observation and data association. Finally, this low accuracy adversely affects vehicle localization.

The RMSE and maximum absolute error (MAE) of vehicle position and the execution time of methods during the path (described in Fig. \ref{fig:fig_10}) are presented in Table. \ref{TB_2}, for the EKF, UnFS, ICKF, and LMKF methods. It can be seen that the LMKF despite of imprecise estimation of robot angle sensor measurements, has a better performance and lower error in comparison to the other methods.

\begin{table}[h]
    \centering
    \caption{RMSE and MAE of vehicle position and execution time during the path described in \ref{fig:fig_10} for EKF, UnFS, ICKF, and LMKF methods.}
    \label{TB_2}
    \begin{tabular}{p{1.2cm} p{2.5cm} p{2.5cm} p{2.5cm}}
        \hline
        & position RMSE & position MAE & execution time \\
        \hline
        EKF   & 2.58 (m) & 4.41 (m) & 52.56 (sec) \\
        UnFS  & 2.1 (m)  & 3.92 (m) & 92.88 (sec) \\
        ICKF  & 2.33 (m) & 4.2 (m)  & 70.66 (sec) \\
        LMKF  & \textbf{1.2 (m)} & \textbf{3.46 (m)} & \textbf{45.43} (sec) \\
        \hline
    \end{tabular}
\end{table}

Note that in the Victoria Park dataset, wheelbase $D\ $is equal to $2.83m$, the sampling time $d_t$ is equal to $25ms$ and the average of robot speed along the described path in Fig. \ref{fig:fig_10} is $2.3m/s$.

\section{Conclusions}\label{sec4}

In this article, we showed that by applying a simple but effective transformation to linearize the state space model, the SLAM could be implemented with better accuracy and performance. The proposed method converged and had a lower error in comparison with the well-known SLAM methods. It was shown that the new approach was robust against changes in system parameters and robot sensors uncertainties, while these uncertainties had significant effects on EKF, UnFS, and ICKF performance.

In all simulation tests, the sensor values were chosen such that the simulation conditions of EKF, UnFS, and ICKF would be better than those of LMKF. Repeated simulations in the same conditions showed that the variances of the robot and landmarks positions errors in LMKF were less than the other methods that show the robustness of the proposed method to different noise sequences. In common methods, the positioning error in the open-loop path was highly more than that in the closed-loop path, while, in the new method this difference reached a maximum of 20 percent. In all of the performed tests, even in unfair conditions in favor of competitors, the stability and performance of the proposed method were better than the other SLAMs.

\bibliographystyle{plain}
\bibliography{mybibliography}

\begin{thebibliography}{10}

\bibitem{Bailey2004}
T.~Bailey.
\newblock {\em Mobile robot localisation and mapping in extensive outdoor
  environments}.
\newblock PhD thesis, University of Sydney, 2004.

\bibitem{Bailey2006}
T.~Bailey, J.~Nieto, J.~Guivant, M.~Stevens, and E.~Nebot.
\newblock Consistency of the ekf-slam algorithm.
\newblock In {\em Proceedings of the IEEE/RSJ International Conference on
  Intelligent Robots and Systems}, pages 3562--3568, 2006.

\bibitem{Barrau2014}
A.~Barrau and S.~Bonnabel.
\newblock The invariant extended kalman filter as a standard observer.
\newblock In {\em Proceedings of the IEEE Conference on Decision and Control},
  pages 1023--1028, 2014.

\bibitem{Barrau2015}
A.~Barrau and S.~Bonnabel.
\newblock An ekf-slam algorithm with consistency properties.
\newblock {\em IEEE Transactions on Robotics}, 31(1):1--15, 2015.

\bibitem{Castellanos2007}
J.~A. Castellanos, R.~Martinez-Cantin, J.~D. Tardos, and J.~Neira.
\newblock Robocentric map joining: Improving the consistency of ekf-slam.
\newblock {\em Robotics and Autonomous Systems}, 55(1):21--29, 2007.

\bibitem{Castellanos2004}
J.~A. Castellanos, J.~Neira, and J.~D. Tardos.
\newblock Limits to the consistency of ekf-based slam.
\newblock In {\em Proceedings of the 5th IFAC Symposium on Intelligent
  Autonomous Vehicles}, 2004.

\bibitem{Chandra2011}
K.~P.~B. Chandra and D.~Gu.
\newblock Cubature kalman filter based slam.
\newblock In {\em Proceedings of the IEEE International Conference on Robotics
  and Automation}, pages 1--6, 2011.

\bibitem{Dissanayake2011}
G.~Dissanayake, S.~Huang, Z.~Wang, and R.~Ranasinghe.
\newblock A review of recent developments in simultaneous localization and
  mapping.
\newblock In {\em Proceedings of the 6th International Conference on Industrial
  and Information Systems}, pages 477--482, 2011.

\bibitem{Dissanayake2001}
G.~Dissanayake, P.~Newman, S.~Clark, H.~F. Durrant-Whyte, and M.~Csorba.
\newblock A solution to the simultaneous localization and map building (slam)
  problem.
\newblock {\em IEEE Transactions on Robotics and Automation}, 17(3):229--241,
  2001.

\bibitem{Durrant2006}
H.~Durrant-Whyte and T.~Bailey.
\newblock Simultaneous localization and mapping: part i.
\newblock {\em IEEE Robotics \& Automation Magazine}, 13(2):99--110, 2006.

\bibitem{Grisetti2010}
G.~Grisetti, R.~Kummerle, C.~Stachniss, and W.~Burgard.
\newblock A tutorial on graph-based slam.
\newblock {\em IEEE Intelligent Transportation Systems Magazine}, 2(4):31--43,
  2010.

\bibitem{Ho2015}
H.~W. Ho and S.~K. Yeung.
\newblock A review of simultaneous localization and mapping.
\newblock {\em International Journal of Advanced Robotic Systems}, 12(12):175,
  2015.

\bibitem{Huang2009}
G.~P. Huang, A.~I. Mourikis, and S.~I. Roumeliotis.
\newblock A first-estimates jacobian ekf for improving slam consistency.
\newblock In {\em Proceedings of the International Symposium on Experimental
  Robotics}, pages 1--10, 2009.

\bibitem{Huang2010}
G.~P. Huang, A.~I. Mourikis, and S.~I. Roumeliotis.
\newblock Observability-based rules for designing consistent ekf slam
  estimators.
\newblock {\em The International Journal of Robotics Research}, 29(5):502--528,
  2010.

\bibitem{Huang2006}
S.~Huang and G.~Dissanayake.
\newblock Convergence and consistency analysis for extended kalman filter based
  slam.
\newblock {\em IEEE Transactions on Robotics}, 23(5):1036--1049, 2006.

\bibitem{Huang2008}
S.~Huang and G.~Dissanayake.
\newblock On the observability and the consistency of the ekf-slam.
\newblock {\em IEEE Transactions on Robotics}, 24(4):901--908, 2008.

\bibitem{Huang2007}
S.~Huang, G.~Dissanayake, and Z.~Wang.
\newblock A novel approach to the convergence and consistency analysis of
  ekf-slam.
\newblock In {\em Proceedings of the IEEE International Conference on Robotics
  and Automation}, pages 2210--2215, 2007.

\bibitem{Hui2009}
Y.~Hui and Z.~Bing.
\newblock Research on consistency of ekf-slam.
\newblock In {\em Proceedings of the IEEE International Conference on
  Information and Automation}, pages 876--881, 2009.

\bibitem{Julier2001}
S.~J. Julier and J.~K. Uhlmann.
\newblock A counter example to the theory of simultaneous localization and map
  building.
\newblock In {\em Proceedings of the IEEE International Conference on Robotics
  and Automation}, pages 4238--4243, 2001.

\bibitem{Kim2008}
C.~Kim, R.~Sakthivel, and W.~K. Chung.
\newblock Unscented fastslam: A robust algorithm for the simultaneous
  localization and mapping problem.
\newblock In {\em Proceedings of the IEEE International Conference on Robotics
  and Automation}, pages 2439--2445, 2008.

\bibitem{Kummerle2011}
R.~Kummerle, G.~Grisetti, H.~Strasdat, K.~Konolige, and W.~Burgard.
\newblock g2o: A general framework for graph optimization.
\newblock In {\em Proceedings of the IEEE International Conference on Robotics
  and Automation}, pages 3607--3613, 2011.

\bibitem{Lee2011}
S.~Lee and J.~Choi.
\newblock The effect of measurement noise on the consistency of ekf-slam.
\newblock In {\em Proceedings of the IEEE International Conference on Robotics
  and Automation}, pages 1--6, 2011.

\bibitem{Mourikis2004}
A.~I. Mourikis and S.~I. Roumeliotis.
\newblock A dual-layer estimator architecture for long-term localization.
\newblock In {\em Proceedings of the IEEE International Conference on Robotics
  and Automation}, pages 2529--2536, 2004.

\bibitem{thrun2004}
S.~Thrun and J.~J. Leonard.
\newblock Simultaneous localization and mapping.
\newblock In {\em Springer Handbook of Robotics}, pages 871--889. 2004.

\bibitem{Zhao2013}
L.~Zhao, S.~Huang, Y.~Sun, and G.~Dissanayake.
\newblock A new approach to graph-based slam.
\newblock In {\em Proceedings of the IEEE International Conference on Robotics
  and Automation}, pages 3182--3187, 2013.

\end{thebibliography}
\end{document}